%% file: paper.tex
\documentclass[11pt]{article}
\usepackage{coling2018}
\usepackage{times}
\usepackage{url}
\usepackage{latexsym}
\usepackage{float}
\usepackage{listings}

\usepackage{latexsym,multirow,graphicx,graphbox}
\usepackage[export]{adjustbox}
\usepackage{epstopdf,comment}
\usepackage{url}

\title{Transparent, Efficient, and Robust\\Word Embedding Access with WOMBAT}

\author{Mark-Christoph M\"uller\and Michael Strube\\
  Heidelberg Institute for Theoretical Studies gGmbH\\
  Schloss-Wolfsbrunnenweg 35\\
  69118 Heidelberg, Germany\\
  {\tt \{mark-christoph.mueller|michael.strube\}@h-its.org}}

\date{}

\begin{document}
\maketitle
\begin{abstract}
We present WOMBAT, a Python tool which supports NLP practitioners in accessing word embeddings from code. WOMBAT addresses common research problems, including unified access, scaling, and robust and reproducible preprocessing. Code that uses WOMBAT for accessing word embeddings is not only cleaner, more readable, and easier to reuse, but also much more efficient than code using standard in-memory methods: a Python script using WOMBAT for evaluating seven large word embedding collections ($8.7$M embedding vectors in total) on a simple SemEval sentence similarity task involving $250$ raw sentence pairs completes in under ten seconds end-to-end on a standard notebook computer.
\end{abstract}
\input{introduction.tex}
\input{wombat.tex}

\input{usecase.tex}
\input{acknowledgements.tex}

\bibliographystyle{acl}
\bibliography{../../bib/extra_lit,../../bib/lit,xabstract}
\end{document}

%% file: introduction.tex
\section{Motivation}
\label{sec:introduction}
\blfootnote{
     \hspace{-0.65cm}  
     This work is licensed under a Creative Commons 
     Attribution 4.0 International License.
     License details:
     \url{http://creativecommons.org/licenses/by/4.0/}}
Word embeddings are ubiquitous resources in current NLP which normally come as plain-text files containing collections of \texttt{$<$string, real-valued vector$>$} tuples.
Each word embedding collection (WEC) is uniquely identified by its combination of 1) training \emph{algorithm}, 2) training \emph{parameters}, and 3) training \emph{data}. 
The latter, in turn, is characterized by the \emph{textual raw data} and the \emph{preprocessing} that was applied to it.

\noindent
Word embeddings are often used early on in the system pipeline: in a typical setup, a word embedding file is loaded up-front (\emph{eager loading}), and vectors are looked up in memory and used as replacements for 
input words. 
This \emph{native} approach to word embedding access has a couple of limitations with respect to transparency, efficiency, and robustness.

1.\ Writing code in which WECs are \textbf{easily and unambiguously identified} is difficult when each WEC is treated as a monolith in the file system.
This way of identifying WECs completely disregards  -- and, in the worst case, obscures -- the fact that these resources might \emph{share} some of their meta data, resulting in different degrees of similarity between WECs: two or more WECs might be identical except for their training window sizes, or except for the fact that some additional postprocessing was applied to one of them. 
For intrinsic and extrinsic evaluation \cite{schnabel2015,nayak2016} of the effect of different training parameters on WECs, these parameters need to be accessible explicitly, and not just on the level of file names.

2.\ Experiments with \textbf{large numbers of WECs} do not scale and are inefficient if entire files need to be loaded every time.
Experiments involving large numbers of WECs are not uncommon: \newcite{baroni2014} employed $48$ different WECs, while \newcite{levy2015} used as many as $672$. More recently, \newcite{wendlandt2018} explored the (in)stability of word embeddings by evaluating WECs trained for all combinations of three algorithms (two of them involving a random component), five vector sizes (dimensions), and seven data sets. 
In order to include the effect of randomness, five sets of WECs with different initializations were trained for the two algorithms, resulting in $385$ WECs altogether. 
\newcite{antoniak2018} focused on training \emph{corpora}, in particular on the effect of three different sampling methods. They trained WECs for all combinations of these three methods, four algorithms, six data sets, and two segmentation sizes. To tackle the effect of randomness, they trained repeatedly for $50$ times, producing a total of $7.200$ WECs. 
None of these papers provides technical details on how WECs are handled, but the code that is available indicates that the native, eager loading approach seems to be prevalent.
More sophisticated, selective access to stored WECs is required to 
speed up experimentation and also support more ad-hoc, explorative approaches. 

3.\ Finally, \textbf{converting unrestricted raw data into units for WEC vector lookup} often amounts to guesswork because the original preprocessing code is not shared together with the WEC resource.
Preprocessing -- which can involve everything from lowercasing, tokenization, stemming or lemmatization, to stop word and special character removal, right up to detecting and joining multiword expressions -- is often underestimated in NLP, and word embedding research is not an exception: 
For the well-known and widely used GloVe embeddings, the documentation simply states to first use "something like the Stanford Tokenizer"\footnote{\url{https://github.com/stanfordnlp/GloVe/blob/master/src/README.md}}.
The $100$B word data set used to train the GoogleNews embeddings\footnote{\url{https://drive.google.com/file/d/0B7XkCwpI5KDYNlNUTTlSS21pQmM/edit?usp=sharing}} contains a considerable number of automatically detected multiword expressions. As a result, as many as $2.070.978$ of the $3$M vocabulary items are phrases joined with one or more "\_" characters. Standard preprocessing without access to the same phrase extraction code cannot detect these items and will thus be blind to almost $70 \%$ of the GoogleNews WEC vocabulary. 
Any preprocessing code used in the creation of a WEC resource has to be considered an integral part of that resource. This is the only way to ensure that the resource is fully (re)usable, which in turn is a prerequisite for the reproducibility of experiments utilizing that resource. 
The topic of \emph{reproducibility} has been around in e.g.\ computational biology for some time \cite{sandve2013}, and is also gaining attention in NLP (see e.g.\ the 4REAL workshops in 2016 and 2018).
Already in 2013, Fokkens et al.\ identified preprocessing, in particular tokenisation, as one of the major sources of errors in their attempts to reproduce NER results.\footnote{\newcite{fokkens2013} do not address preprocessing for word embeddings, but their conclusions apply just the same.}

\noindent
While some word embedding APIs and toolkits do exist,\footnote{E.g.\ \url{https://radimrehurek.com/gensim/models/word2vec.html}, \url{https://github.com/3Top/word2vec-api}, \url{https://github.com/stephantul/reach}, \url{https://github.com/vecto-ai/vecto}} they mostly focus on providing interfaces for in-memory vector lookup or for higher-level similarity tasks. None of them adresses scalability or preprocessing issues. 

%% file: wombat.tex
\section{WOMBAT in a Nut Shell}
\label{sec:wombat}
\textbf{WOMBAT}, the \textbf{WO}rd e\textbf{MB}edding d\textbf{AT}abase, is a light-weight Python tool for more transparent, efficient, and robust access to potentially large numbers of WECs. 
It supports NLP researchers and practitioners in developing compact, efficient, and reusable code.
Key features of WOMBAT are
1.\ \textbf{transparent} identification of WECs by means of a clean syntax and human-readable features,
2.\ \textbf{efficient} \emph{lazy}, on-demand retrieval of word vectors, and
3.\ increased \textbf{robustness} by systematic integration of executable preprocessing code. 
WOMBAT implements some \emph{Best Practices} for research reproducibility \cite{sandve2013,stodden2014}, and complements existing approaches towards WEC standardization and sharing.\footnote{E.g.\ \url{http://vectors.nlpl.eu/repository}, \url{http://wordvectors.org}, \url{https://github.com/JaredFern/VecShare}, \url{http://bit.ly/embeddings}}
The WOMBAT source code including sample WEC data is available
at \url{https://github.com/nlpAThits/WOMBAT}.

\noindent
WOMBAT
provides a single point of access to \emph{existing} WECs.
Each plain text WEC file has to be imported into WOMBAT \emph{once}, receiving in the process a set of \texttt{ATT:VAL} identifiers consisting of five system attributes (\texttt{algo, dims, dataset, unit, fold}) plus arbitrarily many user-defined ones. 

\begin{scriptsize}
\begin{lstlisting}[frame=single, title=Importing the GoogleNews embeddings into WOMBAT., captionpos=b, basicstyle=\ttfamily]
from wombat import connector as wb_conn
wbc = wb_conn(path="/home/user/WOMBAT-data/", create_if_missing=True)
wbc.import_from_file("GoogleNews-vectors-negative300.txt",
                     "algo:w2v;dataset:googlenews;dims:300;fold:0;unit:token")
\end{lstlisting}
\end{scriptsize}
The above code is sufficient to import the GoogleNews embeddings. The combination of identifiers, provided as a semicolon-separated string, must be unique, but the supplied order is irrelevant. In this example, no additional user-defined attributes were assigned, as the publicly available GoogleNews WEC is sufficiently identifiable. 
For self-trained WECs, user-defined attributes for hyper-parameters including minimum frequency, window size, and training iterations are usually employed. 

\noindent
In WOMBAT, each WEC is stored in a single one-table relational database\footnote{We currently use SQLite (\url{https://sqlite.org}).} with a \emph{word} column and a \emph{vector} column as a \texttt{float32} NumPy array, which significantly reduces the disk size, 
e.g.\ from $12.7$ GB to $4.1$ GB for GoogleNews.
In order to maintain data integrity, the \emph{word} column employs a \texttt{unique} database index to prevent multiple entries for the same word.

\noindent
The most basic WOMBAT operation is the retrieval of embedding vectors from one or more WECs, which are specified by their identifiers. 
For this, WOMBAT provides a \texttt{get\_vectors(...)} method supporting a grid search-friendly \texttt{ATT:\{VAL1,VAL2,...,VALn\}} identifier format, which is expanded into \texttt{n} atomic identifiers in the supplied order. In addition, several WEC identifiers can be concatenated with \texttt{\&}. If input words are already preprocessed, they can directly be supplied as a nested Python \texttt{list}. 

\noindent
The following code retrieves vectors for the words \texttt{theory} and \texttt{computation} from all six GloVe WECs and from the GoogleNews WEC, in under two seconds on a standard notebook computer. 
The special identifier format is used here to specify all four GloVe 6B data sets, which share all properties except for \texttt{dims} (vector dimensionality). Other typical uses supported by this format include the evaluation of WECs trained with different hyper-parameters like e.g.\ \texttt{window}. 

\begin{scriptsize}
\begin{lstlisting}[frame=single, title=Retrieving embedding vectors from several WECs., captionpos=b, basicstyle=\ttfamily]
from wombat import connector as wb_conn
wbc = wb_conn(path="/home/user/WOMBAT-data/")
wecs = "algo:glove;dataset:6b;dims:{50,100,200,300};fold:1;unit:token&\
        algo:glove;dataset:42b;dims:300;fold:1;unit:token&\
        algo:glove;dataset:840b;dims:300;fold:0;unit:token&\
        algo:w2v;dataset:googlenews;dims:300;fold:0;unit:token"
v = wbc.get_vectors(wecs, {}, for_input=[["theory", "computation"]], raw=False, as_tuple=True)
\end{lstlisting}
\end{scriptsize}

\noindent
More often, however, input text is \emph{raw} and needs to be preprocessed into smaller units before word vectors can be retrieved. 
WOMBAT acknowledges the importance of preprocessing by providing a two-level mechanism for directly integrating user-defined preprocessing code. The first, obligatory level handles the actual preprocessing by piping each raw input line through a \texttt{process(...)} method. User-defined Python code implementing this method is directly inserted into the WOMBAT database. When vectors for raw input text are to be retrieved (\texttt{raw=True}), the correct preprocessing for each WEC is automatically applied in the background.
While each WEC in WOMBAT could have its own preprocessing, the expected input format for many WECs (e.g.\ GloVe) is almost identical. Only \texttt{glove.840B.300d.txt}, e.g., is case-sensitive, while the others are not. This difference, encoded in the WOMBAT meta data as \texttt{fold:0} and \texttt{fold:1}, is accounted for automatically during preprocessing. 
Similarly, some WECs might exist in both an unstemmed and a stemmed variant, which can be distinguished by the values \texttt{token} and \texttt{stem} in the \texttt{unit} attribute. These values are also evaluated during preprocessing.
The second, optional processing level analyses the token sequence produced by the first level and joins into phrases those adjacent tokens for which vocabulary items exist in the WEC. Currently this is done by a \texttt{gensim.models.phrases.Phraser} object, which initially needs to be trained on the tokenized textual raw data \emph{before} WEC training, and which then needs to be applied to this data in order to enrich it with phrase information.

\noindent
WOMBATs \texttt{get\_vectors(...)} method returns data as a generic, nested Python data structure.
Basically, it is a list containing one two-item tuple for every WEC, where the first item is the WEC identifier, and the second item is a nested structure containing the actual result, including the raw and preprocessed input and a list of \texttt{$<$word, vector$>$} tuples. By default, the ordering of this tuple list is undefined, but input ordering can optionally be maintained (\texttt{in\_order=True}).
For most tasks, however, ordering is irrelevant, which is why the more efficient \texttt{in\_order=False} is the default.

\vspace{22pt}

\begin{tiny}
\begin{lstlisting}[frame=single, title=Schematic WOMBAT result format., captionpos=b, basicstyle=\ttfamily]
[                                                     # top-level result container
 [                                                    # start of result for first WEC
  (
   'algo:glove;dataset:6b;dims:50;fold:1;unit:token', # normalized WEC identifier
   [
    (
     [],                                              # raw input as supplied to for_input (empty here since raw=False)
     ['theory','computation'],                        # tokens produced by preprocessing (if raw=True)                                                      
     [                                                    
      (                                               # result tuple for 'theory'
       'theory',                                      # token exactly as used in lookup
       [0.28217, 0.65819001, ... -0.39082, -0.1266]   # vector as NumPy array
      ),             
      (                                               # result tuple for 'computation'
       'computation',                                 # token exactly as used in lookup
       [-0.25176001, -0.028599, ... 0.31508, 0.25172] # vector as NumPy array
      )           
     ]
    )                                                   
   ]                                                  # end of result for first (and only) input list          
  )
 ],                                                   # end of result for first (and only) WEC
 ...                                                  # potential result for second WEC 
]
\end{lstlisting}
\end{tiny}

%% file: usecase.tex
\section{Sample Use Cases}
\label{sec:usecase}
At this point, the \texttt{wombat.analyse} library contains only a few methods (cf.\ below). Our focus has been on developing a stable, generic, and efficient code base, on top of which more complex and useful functionality (incl.\ further visualizations, nearest neighbors, etc.) can be implemented.

\subsection{Global Sentence Similarity}
\label{sec:usecaseglobalsentencesimilarity}
In order to demonstrate WOMBAT in an actual end-to-end use case, we applied it to a sentence pair similarity ranking task, using the data set from task 1, track 5 of SemEval-2017 \cite{cer2017}. The data set consists of $250$ tab-separated, raw sentence pairs. 
Since we focus on preprocessing and vector retrieval, we implement a simple baseline approach only, in which sentences are represented as the average vector of their respective word vectors (excluding stop words) and the pairwise distances are computed as cosine distance. The result is a list containing, for each WEC, an ordered list of tuples of the form \texttt{$<$distance, sentence1, sentence2$>$}. The following code implements the whole process. 
The distance metric in the \texttt{pairwise\_distances(...)} method is provided as a parameter, and can be set to any method for computing vector distances (or similarities, in which case the output ordering can be reversed with \texttt{reverse=True}).

\begin{tiny}
\begin{lstlisting}[frame=single, title=Global sentence similarity computation with WOMBAT., captionpos=b, basicstyle=\ttfamily]
import numpy, scipy.spatial.distance
from wombat import connector as wb_conn
from wombat.analyse import pairwise_distances
wbc = wb_conn(path="/home/user/WOMBAT-data/")
wecs = "algo:glove;dataset:6b;dims:{50,100,200,300};folded:1;unit:token&algo:glove;dataset:42b;dims:300;folded:1;unit:token&\
        algo:glove;dataset:840b;dims:300;folded:0;unit:token&algo:w2v;dataset:googlenews;dims:300;folded:0;unit:token"
infile = "STS.input.track5.en-en.txt"
pp_cache = {}
vecs_1 = wb.get_vectors(wecs, pp_cache, for_input=[numpy.loadtxt(infile, dtype=str, delimiter='\t', usecols=0)], raw=True) 
vecs_2 = wb.get_vectors(wecs, pp_cache, for_input=[numpy.loadtxt(infile, dtype=str, delimiter='\t', usecols=1)], raw=True) 
pd = pairwise_distances(vecs_1, vecs_2, metric=scipy.spatial.distance.cosine, reverse=False)
\end{lstlisting}
\end{tiny}
\noindent
The execution time for reading the input file (column 0 and column 1 separately), preprocessing, vector retrieval from seven WECs, vector averaging per sentence, pairwise distance computation, and sorting is under ten seconds on a standard notebook computer. 

\subsection{Word-Level Sentence Similarity}
\label{sec:usecasedblpsentencesimilarity}
\begin{figure}[H]
\centering
\includegraphics[align=c, width=0.48\textwidth, trim=8 32 60 10,clip]{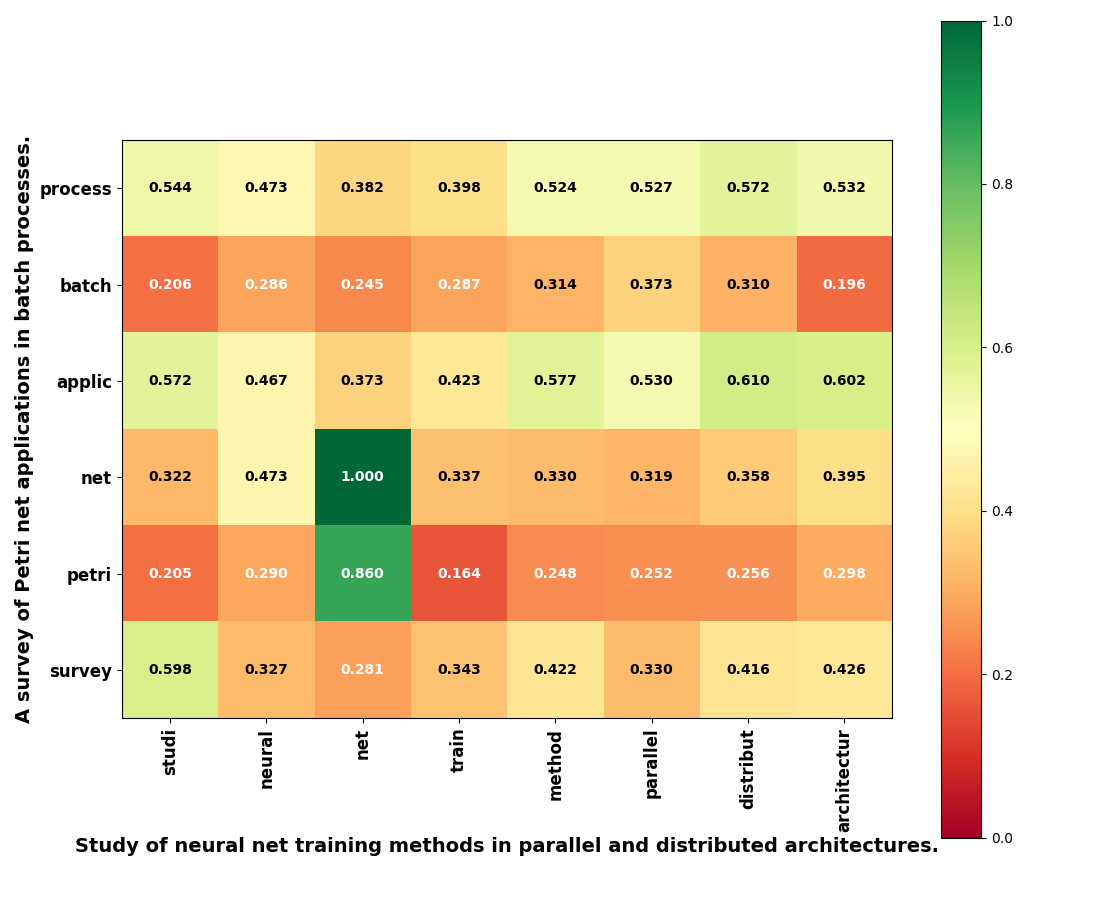}                
\includegraphics[align=c, width=0.48\textwidth, trim=8 32 60 10,clip]{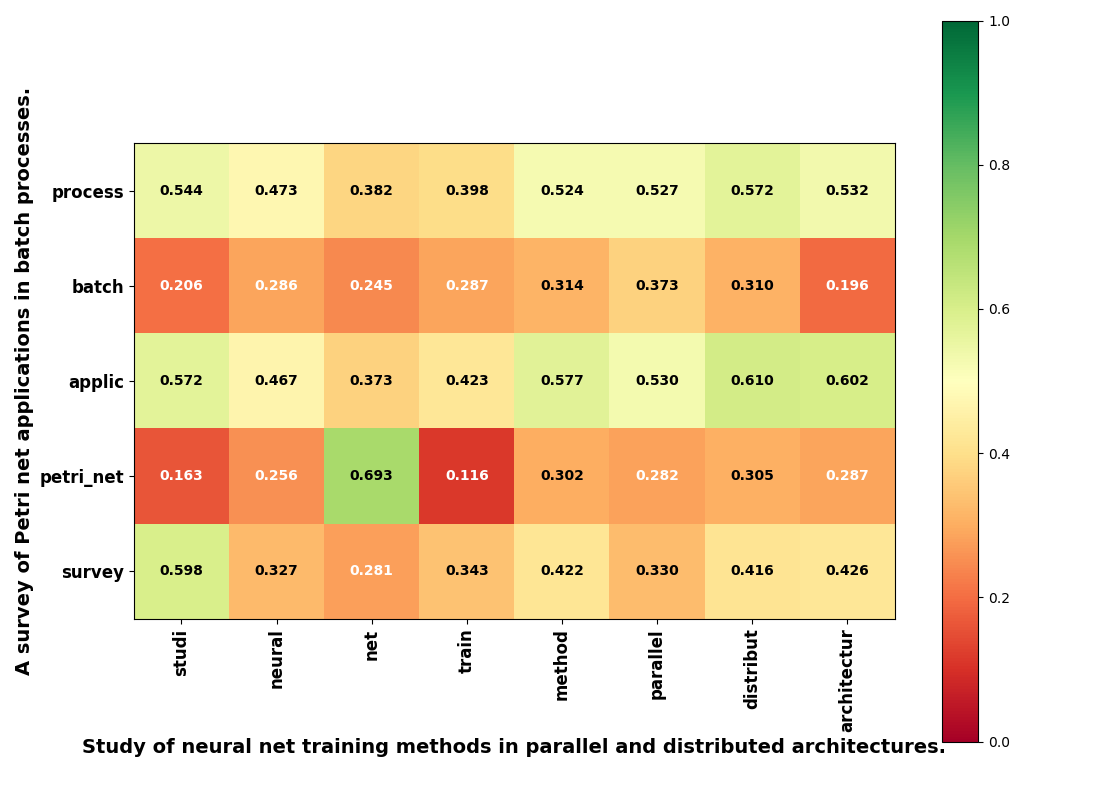}                
\caption{Word-level sentence similarity without (left) and with (right) phrase-aware preprocessing.}
\label{fig:petri}
\end{figure}

WOMBAT was originally developed in a research project dealing with scientific publication title similarity, which involved light-weight semantic matching based on WEC-based similarities. Figure \ref{fig:petri} shows two sample outputs of WOMBATs \texttt{plot\_heatmap(...)} method, which accepts as input the generic output vectors produced by \texttt{get\_vectors(...)}. 
The two plots show the contribution of phrase-aware preprocessing in the comparison of two publication title strings: the left plot was fed with \texttt{$<$string, vector$>$} tuples which were created with phrases temporarily disabled, and shows a spurious maximal similarity for the term \emph{net} in the two titles.
The right plot, in contrast, was fed with tuples which were created with phrases enabled, including a separate vector for \emph{Petri net}.
The plot shows a more differentiated, still high, but not maximal similarity between \emph{Petri net} and \emph{net}, resulting in a more accurate general representation of the to titles' similarities. Heat maps, of course, are standard visualization, but WOMBAT provides methods for their efficient, large-scale creation.  

%% file: acknowledgements.tex
\noindent
\textbf{Acknowledgements}
\noindent
The research described in this paper was conducted in the project \emph{SCAD -- Scalable Author Name Disambiguation}, funded in part by the Leibniz Association (grant SAW-2015-LZI-2), and in part by the Klaus Tschira Foundation. We thank the anonymous COLING reviewers for their useful suggestions.